\documentclass[10pt,twocolumn,letterpaper]{article}

\usepackage{iccv}
\usepackage{times}
\usepackage{epsfig}
\usepackage{graphicx}
\usepackage{amsmath}
\usepackage{amssymb}
\usepackage{booktabs}
\usepackage{xcolor}
\usepackage{adjustbox}
\usepackage{makecell}
\usepackage{multirow}
\usepackage{bm}
\usepackage{verbatim}
\usepackage{enumitem} 
\usepackage{arydshln}
\usepackage{subfigure}

\DeclareMathOperator*{\argmax}{arg\,max}

\def\a{\mathbf{a}}
\def\bo{\mathbf{o}} 

\def\x{\mathbf{x}}

\def\1{\mathbf{1}}

\def\X{\mathbf{X}}

\def\cA{\mathcal{A}} 
\def\cO{\mathcal{O}} 
\def\cX{\mathcal{X}}

\usepackage[pagebackref=true,breaklinks=true,colorlinks,bookmarks=false]{hyperref}

\iccvfinalcopy 

\def\iccvPaperID{6649} 

\ificcvfinal\fi

\begin{document}

\title{The Spatio-Temporal Poisson Point Process: \\ A Simple Model for the Alignment of Event Camera Data}

\author{Cheng Gu\\
TU Berlin\\
{\tt\small c.gu@campus.tu-berlin.de}
\and
Erik Learned-Miller\\
UMass Amherst\\
{\tt\small elm@cs.umass.edu}
\and
Daniel Sheldon\\
UMass Amherst\\
{\tt\small sheldon@cs.umass.edu}
\and
Guillermo Gallego\\
TU Berlin\\
{\tt\small guillermo.gallego@tu-berlin.de}
\and
Pia Bideau\\
TU Berlin\\
{\tt\small p.bideau@tu-berlin.de}
}

\maketitle
\ificcvfinal\thispagestyle{empty}\fi

\begin{abstract}
Event cameras, inspired by biological vision systems, provide a natural and data efficient representation of visual information. Visual information is acquired in the form of events that are triggered by local brightness changes. Each pixel location of the camera's sensor records events asynchronously and independently with very high temporal resolution. However, because most brightness changes are triggered by relative motion of the camera and the scene, the events recorded at a single sensor location seldom correspond to the same world point. 
To extract meaningful information from event cameras, it is helpful to register events that were triggered by the same underlying world point.
In this work we propose a new model of event data that captures its natural spatio-temporal structure. We start by developing a model for {\bf {\em aligned}} event data. That is, we develop a model for the data as though it has been perfectly registered already. In particular, we model the aligned data as a {\bf {\em spatio-temporal Poisson point process}}. Based on this model, we develop a maximum likelihood approach to registering events that are not yet aligned. That is, we find transformations of the observed events that make them as likely as possible under our model. In particular we extract the camera rotation that leads to the best event alignment. We show new state of the art accuracy for rotational velocity estimation on the DAVIS 240C dataset~\cite{Mueggler17ijrr}. In addition, our method is also faster and has lower computational complexity than several competing methods.
Code: \url{https://github.com/pbideau/Event-ST-PPP}
\end{abstract}

\section{Introduction}
\label{sec:intro}

\begin{figure}
\begin{center}
\subfigure[Method overview]
{
    \includegraphics[width=0.95\columnwidth]{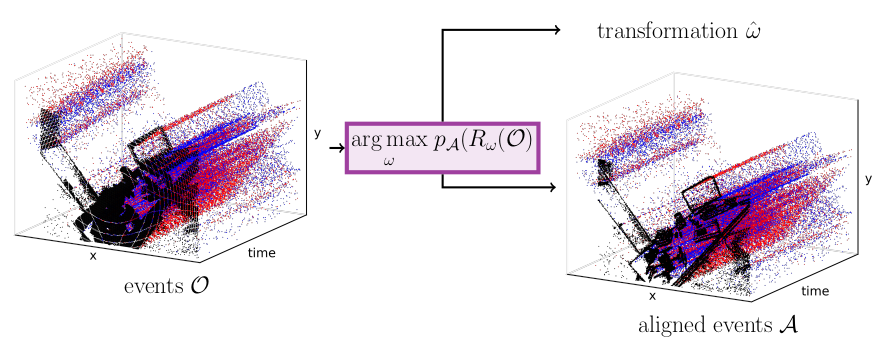}
}\\
\subfigure[Video frame]
{
    \includegraphics[width=0.28\columnwidth]{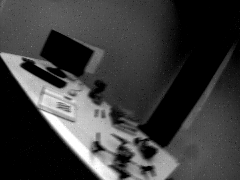}
}
\subfigure[Unaligned events - "blurred" event image]
{
    \includegraphics[width=0.28\columnwidth]{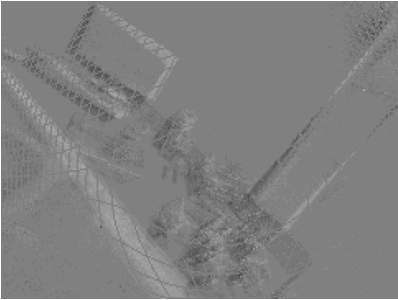}
    \label{fig:events-blurred}
}
\subfigure[Aligned events - sharp event image]
{
    \includegraphics[width=0.34\columnwidth]{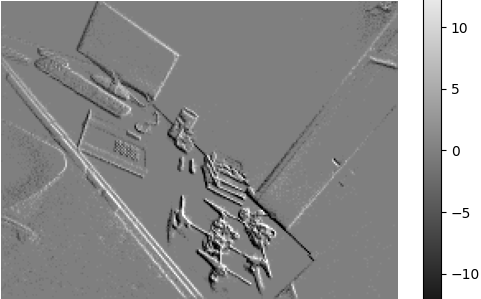}
    \label{fig:events-aligned}
}
\caption[Overview]{Alignment of event data by maximizing the joint probability of a set of events $p_\cA(R_\omega(\cO))$. 
Top row: Events are plotted in red/blue depending on their polarity. 
The projection of events onto the 2D image plane at time stamp zero is shown in black - indicating the quality of their alignment over time. 
Bottom row: video frame, accumulated events, accumulated aligned events. 
}
\label{fig:overview}
\end{center}
\end{figure}

Inspired by biological vision systems, event cameras~\cite{Lichtsteiner08ssc,Posch11ssc,Suh20iscas} mimic certain biological features of the human vision system, such as recording brightness changes as \textit{events}, asynchronously, and at high temporal resolution. This relatively new way of acquiring visual information differs significantly from classical frame-based video recordings, leading to new research directions in computer vision and drawing close connections to robotics and the cognitive sciences. 
Prior work has shown that event data is rich enough to recover high quality brightness images, even in high-speed and high dynamic range (HDR) scenarios~\cite{Rebecq19pami}, and it allows early stage information processing such as motion perception and recognition~\cite{Zhu18rss,Lagorce17pami}.
 Despite these advantages, current vision algorithms still struggle to unlock the benefits of events cameras. 


{\bf The problem of aligning event camera data.}
In this paper we focus on event camera data that comes from a moving camera in a static or nearly static environment. Because of the camera motion, as the camera records events through time, the events at a fixed camera pixel correspond to different points in the world. 
This makes it more difficult to interpret event camera data, since accumulating events through time corresponds to mixing events from different world point sources. Finding transformations of the events that map each event triggered by the same world point to the same pixel location of the camera sensor can be called {\em alignment} or {\em registration} of the events. In this paper, we propose a method for alignment based on a new probabilistic model for event camera data. 

{\bf Panoramas of events.}
To describe our model and algorithm, we draw analogies with {\em image panoramas} created using RGB images. By warping a set of images taken from different camera positions into the same shared set of coordinates, a set of images may be combined into a larger composite image, or panorama, of a scene. 

The same idea can be applied to event data: transforming the location of each individual event so that it is transformed into a shared coordinate system~\cite{Reinbacher17iccp, Kim14bmvc}.\footnote{In the event camera literature, the term `panorama' is usually applied to alignment over sequences in which the camera has large displacements, resulting in a panorama much larger than a single camera frame. However, the same term can be applied to registering short sequences of event camera data, which creates panoramas only slightly larger than a single frame.} Doing this with event data is challenging, since it is more difficult to establish correspondences in event data than among images. 

Instead, many approaches to registering event camera data are based upon a simple intuitive observation~\cite{Mitrokhin18iros, Stoffregen19iccv,Liu20cvpr,Nunes20eccv,Gallego18cvpr}. If we form an `aggregate' event camera image by simply recording the number of events at each pixel over some period of time, then these aggregate images tend to be sharper when the events are well-aligned (Figure~\ref{fig:events-aligned}), and blurrier when the events are less well aligned (Figure~\ref{fig:events-blurred}). 
Leveraging this observation, one tries to find a set of transformations that maximize the sharpness of the aggregate image. These methods, discussed in detail in the related work section, mostly differ in their definition of defining sharpness, i.e., in their loss functions. 

{\bf Congealing and probabilistic models of alignment.}
In this paper, we introduce a new, more effective method for event alignment. 
It is related to a probabilistic method for aligning traditional images known as {\em congealing}~\cite{LearnedMiller06pami}, which does not use any explicit correspondences. Instead, one measures the degree to which a set of images are {\em jointly aligned}. To measure the quality of the joint image alignment, one considers the {\em entropy} of the set of pixels at each image location. If a location has the same pixel value across all of the images, it has minimum entropy. If it has many different pixel values, it has high entropy. By transforming the images so that the sum of  these pixelwise entropies is minimized, the images naturally move into alignment. Since minimizing entropies is equivalent to maximizing pixel likelihoods under a non-parametric distribution, congealing can also be seen as a \emph{maximum likelihood method} 
 (see~\cite{LearnedMiller06pami} for more details).

{\bf Contributions.} We present a novel probabilistic model for event camera data. It allows us to evaluate the likelihood of the event data captured at a particular event camera pixel. By introducing transformations to move the data into a common coordinate system, we show that by maximizing the likelihood of the data under this model with respect to these transformations, we naturally retrieve a very accurate registration of the event data. That is, we develop a probabilistic, maximum likelihood method of joint alignment of event data. We support this novel approach by providing new state-of-the-art results. We have substantially higher accuracy than recently published methods, and are also among the fastest. In addition, we reassess how evaluations on these \emph{de facto} benchmarks are done, and argue that a new approach is needed.

\begin{figure*}
\begin{center}
    \subfigure[\textbf{Observed events} ]{
        \includegraphics[width=0.33\linewidth]{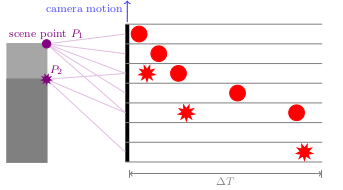}
        \label{fig:irony-a}
    }
    \hspace{3em}
    \subfigure[\textbf{Aligned events}]{
        \includegraphics[width=0.33\linewidth]{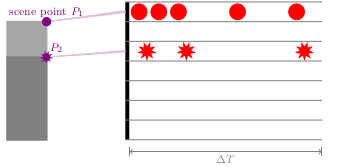}
        \label{fig:irony-b}
    }
   \caption{(a): Camera movement perpendicular to an intensity edge induces brightness changes on the camera sensor plane such that events are generated. Events capturing the same scene point in the world are recorded at different sensor locations - we call them `de-registered'. (b): Events that can be associated with the same scene point in the world are registered to each other and are modeled as aligned Poisson point processes. \emph{Ironically} aligned events are a useful representation of event data to extract scene information, but actually events are only triggered if the camera is moving and thus event-data can only be acquired in its de-registered form (a).}
   \label{fig:irony}
\end{center}
\end{figure*}

\section{Related Work}
\label{sec:relatedWork}

\textbf{Rotational velocity estimation from event data} has been an active research topic since an Inertial Measurement Unit (IMU) was integrated on the Dynamic Vision Sensor (DVS) event camera~\cite{Lichtsteiner08ssc} to yield a combined visual and vestibular device~\cite{Delbruck14iscas} (the precursor of the DAVIS240 event camera~\cite{Brandli14ssc}). 
Sensor fusion between the IMU's gyroscope (measuring angular velocity) and the DVS output allowed the stabilization of events for a short amount of time. 
However, IMUs are interoceptive sensors that suffer from biases and drift (as error accumulate), so exteroceptive solutions using the events were investigated as alternative means to estimate rotational motion and therefore to stabilize event-based output during longer times.

Early work on rotational motion estimation (i.e., camera tracking) from event data includes \cite{Cook11ijcnn,Kim14bmvc,Reinbacher17iccp,Gallego17ral}.
Some of these works arose as \mbox{3-DOF} (degrees of freedom) Simultaneous Localization and Mapping solutions~\cite{Kim14bmvc,Reinbacher17iccp}. 
Since depth cannot be estimated in purely rotational motions, the ``mapping'' part refers to the creation of a panoramic image (either of edges~\cite{Reinbacher17iccp} or of reconstructed brightness~\cite{Kim14bmvc}).
%
The method in~\cite{Gallego17ral} proposed to estimate rotational motion by maximizing the contrast of an image of displaced (warped) events. This contrast measure is the most high quality metric for event alignment in terms of accuracy and computation time among 24 loss functions, that have been explored in~\cite{Gallego19cvpr}.
The event alignment technique has been later applied to other problems (depth, optical flow or homography estimation) in~\cite{Gallego18cvpr}. 
Since then, the idea of event alignment has been gaining popularity and extended, via different alignment metrics and optimization algorithms, for several motion estimation and segmentation problems in~\cite{Gallego19cvpr,Mitrokhin18iros,Stoffregen19iccv,Stoffregen19cvpr,Nunes20eccv,Liu20cvpr,Peng20eccv,Peng21pami,parameshwara20210}.

The closest work to us are~\cite{Nunes20eccv,Gallego18cvpr}.
In \cite{Nunes20eccv} event alignment is expressed via a family of entropy functions over \emph{all pairs} of warped events. 
Entropy measures dispersion and our approach can also be interpreted as an entropy minimization~\cite{LearnedMiller06pami}.
In contrast, we propose a framework that maximizes the likelihood of events at \emph{each pixel location}, as opposed to using pairwise event measures. 
This directly corresponds to minimizing the entropy \emph{per pixel}, independently.
Assuming pixel-wise independence allows to derive an event alignment approach that is computationally efficient (reduced complexity) and achieves high performance, 
as shown in the experiments (Section~\ref{sec:experiments}).
In addition, independent modeling of each pixel leads to a simple theoretical formulation with clear properties and dependencies. 





\textbf{Congealing and probabilistic models for alignment.}
Our event alignment method is inspired by \emph{congealing}~\cite{miller2002learning} --a probabilistic approach for joint alignment of image data. 
Congealing aligns images by maximizing their joint probability under a set of transformations to make the images as similar as possible to each other.
Congealing has been successfully applied to align binary images (e.g., MNIST), medical MRI volumes~\cite{LillaBestPaper}, complex real-world images like faces~\cite{face_congeal,huang2012learning} and 1D curves~\cite{Mattar09icassp}.
In this work, we further develop the principles of congealing to align the unconventional visual data produced by event cameras.
The result is a new probabilistic approach that, while being developed for rotational motion estimation, also extends to related event-alignment problems~\cite{Gallego18cvpr,Nunes20eccv,Peng21pami}.

%

\section{A probabilistic model for event data}
\label{sec:method}
In this section, we present our probabilistic model for event data.
We start by defining two types of `event processes'. These processes are models for the \textit{observed} event data, which is unaligned, and data that has been perfectly aligned using ground truth transformations. These two processes are illustrated in Figure~\ref{fig:irony}.

\subsection{The observed data}

The observed data is a set of $N$ events recorded by a moving event camera over a time period $\Delta T$. We denote the observed events as
\begin{equation}
\cO=[\bo_1,\bo_2,...,\bo_N],
\end{equation}
where $\bo_i=(o_i^\x,o_i^t)$ comprises the pixel location $o_i^\x$ and the time $o_i^t$ at which the event occurred on the image plane.

\subsubsection{The observed pixel processes}
Consider the set of all events recorded at the {\em same pixel} location $\x$ in the event camera. 
Among the events $\cO$, the subset of events $\cO^\x$ that occur at a specific pixel is:
\begin{equation}
\cO^\x=\{\bo_i: o_i^\x=\x\}.
\end{equation}
We refer to such a set of observed events generated at a particular event camera location as an {\em observed pixel process}. Each row of events in Figure~\ref{fig:irony-a} shows such a process. The different shapes in each row illustrate that these pixel processes were generated by different scene points. However, to the camera, the events look the same, irrespective of what world point they were generated from. 

We can define an observed pixel process for each of the $N_P$ pixels in the event camera, resulting in a set of $N_P$ observed pixel processes. We define an observed pixel process for a pixel even if there were no events observed at that pixel. That is, some observed pixel processes may not have any events associated with them.

\subsection{The aligned data}
Next, we consider the events as though they have been perfectly aligned with a set of ideal or ground truth transformations. We describe this as the set of events $$
\cA=[\a_1,\a_2, ..., \a_N].
$$
Here, $\a_i=(a_i^{\x},a_i^t)$ represents an event whose location has been transformed according to $$
a_i^\x = T_{GT}(o_i^\x; t)
$$
where $T_{GT}(\cdot; t)$ is a {\em ground truth} transformation that exactly inverts the camera motion.
We define this ground truth transformation function $T_{GT}$ to be one that maps each event caused by a particular world point $P$ to the  same location in the pixel camera coordinate system. 

This set of aligned events can be thought of as an event panorama in which all of the events have been registered.\footnote{If the camera motion contains translations, then this can only be done approximately.} That is, any events that emanated from the same world point should now have the same coordinates.

Note that like a traditional image panorama, the registration of points is likely to create an `image' which is larger than the original camera image, since we are effectively overlaying a bunch of different images into the same coordinates. Thus, while events are being registered into the same coordinate frame, the actual coordinates may extend beyond the limits of the original image. 

\subsubsection{The aligned pixel processes}
Now for the case of the aligned data, consider the set of all events with the same transformed pixel location $\x$. That is, among the events $\cA$, we define the subset of events $\cA^{\x}$ that occur at a specific pixel location $\x$:
$$
\cA^{\x}=\{\a_i: a_i^{\x}=\x\}.
$$
We refer to such a set of aligned events at a particular  location as the output of an {\em aligned pixel process}. By definition, each event $\bo$ that originates at world point $P$ is transformed to the same image point (with location rounded to the nearest pixel center) $a^\x$ by the ground truth transformation $a^\x=T_{GT}(o^\x;t)$. 
Collectively, the aligned data can be viewed as a set of aligned pixel processes.

\subsection{A probabilistic model for aligned data}
We now introduce our probabilistic model for {\em aligned event camera data} 
and describe how it can be used to align observed (i.e., unaligned) data.

First consider a model for a single aligned pixel process, representing all of the events associated with a particular pixel location $\x$. We model this as a {\em Poisson process}~\cite{Streit2010} with a rate parameter $\lambda_\x$. Henceforth, assume for simplicity time is rescaled so the observation interval length is $\Delta T = 1$. 
This implies that the number of events $k_\x = |\mathcal{A}^\x|$ occurring at location $\x$ is a Poisson random variable with parameter $\lambda_\x$:
\begin{equation}
\label{eq:PoissonOneDataPoint}
    p(k_\x) =  \text{Pois}(k_\x|\lambda_\x) = \frac{\lambda_\x^{k_\x} e^{-\lambda_\x}}{k_\x!}.
\end{equation}


Next, we model the entire aligned data set as the output of a collection of independent Poisson point processes, each with a separate rate $\lambda_\x$ depending upon its location. By standard properties of Poisson processes~\cite{Streit2010}, this is equivalent to a \emph{single} Poisson point process over space and time---i.e., a {\em spatio-temporal Poisson point process} (ST-PPP)---with intensity function $\lambda(\x, t) \doteq \lambda_\x$. By definition, the events of an ST-PPP at one spatial location are independent of those at other locations. 
Fig.~\ref{fig:irony-b} illustrates such an ST-PPP. 


Let $\cX$ be the set of locations at which events occur in the aligned event camera data. Then, due to independence over spatial locations, we can write the probability of the entire aligned data set under the ST-PPP model as\footnote{We slightly abuse notation with the notation $p_\cA(\cA)$; our expression gives the probability of the counts $k_\x$, which differs from the density of the point set $\cA$ by a factor of $k_\x!$.}
\begin{equation}
p_\cA(\cA) \doteq  \label{eq:PoissonProd-1}
\prod_{\x \in \cX} \text{Pois}(k_\x| \lambda_\x), 
\quad k_\x = |\cA^\x|.
\end{equation}

\subsection{An optimization problem}
\label{sec:optim-problem}
The model above leads naturally to an optimization problem. We shall seek a set of transformations, one applied to each event, that maximizes the likelihood of the transformed data under our model. Because the interval $\Delta T$ over which we are considering camera motions is very small (just a fraction of a second), we adopt the typical assumption that our transformations are smooth with respect to time. While we consider other families of transformations in the experiments section,
we describe our optimization with respect to sets of {\em constant angular velocity} rotations:
\begin{eqnarray}\label{eq:rotation}
R_\omega^t = \text{exp}(S(\omega) \cdot t),
\end{eqnarray}
where $S(\omega)$ is a skew-symmetric matrix that encodes the 3-parameter angular velocity $\omega$ and whose exponentiation leads to a rotation matrix. 
Here $t$ is the time of the recorded event, which is set to $0$ for the beginning of the sequence and $\Delta T$ at the end of the sequence. Since $t$ scales the angular velocity $\omega$, it controls the amount of rotation, and hence the amount of rotation is a linear function of $t$.


To transform events, we define $R_\omega^t$ as the mapping $(\x, t) \mapsto (R_\omega^t  \x, t)$ that applies the time-dependent rotation to the event location $\x$ and preserves the event time $t$. In this way, each event is rotated an amount proportional to the time at which it occurred. 

To optimize the alignment of events over this set of choices for transformations, we solve for
\begin{equation}
\hat{\omega}
\,=\,\argmax_{\omega\in\Omega} p_\cA\big(R_\omega(\cO)\big),
\end{equation}
where $R_\omega(\cO) = [R_\omega^{t_0}(\bo_1), \ldots, R_\omega^{\Delta T}(\bo_N)]$. Here, we have implicitly defined the likelihood $p_\cO(\cO | \omega)$ of the observed data through the mapping $p_\cO(\cO | \omega) = p_\cA\big(R_\omega(\cO)\big)$. This can be formally justified through the Poisson mapping theorem~\cite{Streit2010}. We give more background on this in the supplementary material. 

The formula in~\eqref{eq:PoissonProd-1} assumes knowledge of the Poisson rate parameter $\lambda_\x$ at each location $\x$. One option would be to estimate these parameters via maximum likelihood jointly with $\omega$. Instead, we adopt a partially Bayesian approach by maximizing the marginal likelihood of $k_\x$ under the prior $\lambda_\x \sim \text{Gamma}(r, q^{-1}(1-q))$, for fixed parameters $r > 0$ and $q \in [0, 1]$. 
Then, by the well-known construction of the negative binomial distribution as a Gamma-Poisson mixture, the marginal distribution of $k_\x$ is $\text{NB}(r, q)$, which we can compute and optimize directly. Our final model for aligned data, which we will use in place of Eq.~\eqref{eq:PoissonProd-1}, is:
\begin{equation}
p_\cA(\cA) = \prod_{\x \in \X} \text{NB}(k_\x | r, q), \quad k_\x = |\cA^\x|.
\end{equation}
We discuss approaches to estimate the parameters $r$ and $q$ in the experiments section.




\subsubsection{Transformations}
Another choice in event camera alignment algorithms is the choice of transformations. In the most general setting $T(\cdot)$ could be any smooth and invertible map from coordinates $(\x,t)$ to new coordinates $(\x',t)$ describing the new spatial-temporal location of events. Here we focus on camera rotations $R_\omega^t$ as the set of possible transformations, however other transformations such as translations and their combinations are possible. Possible extensions are discussed in the experiments section in further detail.



\subsection{Implementation details}
\label{sec:method:implementation}
\paragraph{Event polarity.}
Until now, we have been considering a single uniform type of event, but most event cameras output either {\em positive} or {\em negative} events depending upon the sign of brightness changes. There are various ways to deal with the diversity of events. One option would be to treat all events as equivalent, irrespective of their polarity, but this would discard information. Instead, we treat positive and negative events as arising from \emph{independent} ST-PPPs--- in other words, the number of positive events $k_\x^+$ and negative events $k_\x^{-}$ at pixel $\x$ in the aligned process are independent Poisson random variables with rates $\lambda_\x^+$ and $\lambda_\x^{-}$, respectively. With $\lambda_\x^+, \lambda_\x^{-} \sim \text{Gamma}(r, q^{-1}(1-q))$, this gives the likelihood:
\begin{equation}
\label{eq:loss:polarity}
p_\cA(\cA) = \prod_{\x \in \cX} \text{NB}(k_\x^+; r, q) \cdot \text{NB}(k_\x^{-}; r, q).
\end{equation}
Operationally, this corresponds to separately computing the log-loss for positive and negative events and adding them together to get a total loss. 


\paragraph{Optimization.} We optimize the loss function~\eqref{eq:loss:polarity} using the Adam algorithm implemented of the python package torch.optim with a learning rate of 0.05 and a maximum number of iterations set to 250. No learning rate decay is applied.
Similar to~\cite{Gallego18cvpr,Gallego17ral} we sequentially process packets of $N=30000$ events, and like~\cite{Gallego17ral} we smooth the image of warped (IWE) events using a Gaussian filter with a small standard deviation ($\sigma = 1$) making the algorithm less susceptible to noise.
We apply a padding of 100 pixels, such that in most cases all recorded events originating from the same world point are aligned with each other and are taken into account for the computation of the loss function. 
The loss function is normalized by the number of events present on the image plane.

\section{Experiments}
\label{sec:experiments}

We evaluate our approach on publicly available data \cite{Mueggler17ijrr}. 
We discuss the results and show an ablation study to support the understanding of our proposed approach for motion estimation from the output of an event camera.
Our approach is based on the event data only and does not require any other additional information such as traditional video frames.

\subsection{Dataset and Evaluation Metrics}
\label{sec:experim:dataset}

The \textbf{DAVIS 240C Dataset} \cite{Mueggler17ijrr} is the de facto standard to evaluate event-camera motion estimation~\cite{Gallego17ral,Liu20cvpr,Nunes20eccv,Reinbacher17iccp,Zhu17cvpr,Rosinol18ral}. 
Each sequence comprises an event stream, video frames, a calibration file, and IMU data from the camera as well as ground truth camera poses from a motion capture system. 
The gyroscope and accelerometer of the IMU output measurements at 1kHz. 
The motion capture system provides ground truth camera poses at 200Hz.
The spatial resolution of the DAVIS camera~\cite{Brandli14ssc} used is $240\times180$ pixels. 
The temporal resolution is in the range of microseconds.
We evaluate our approach on sequences \emph{boxes}, \emph{poster}, \emph{dynamic} and \emph{shapes}. 
All sequences have 1 minute duration, 20--180 million events and an increasing camera motion over time.

\begin{table}
\centering
\small
\begin{adjustbox}{max width=0.95\linewidth}
\setlength{\tabcolsep}{3pt}
\begin{tabular}{p{2ex}lcccccc}
\toprule 
& \bf{Method} & $e_{wx}$ & $e_{wy}$ & $e_{wz}$ & $\sigma_{ew}$ & RMS & RMS\% \\
\midrule 
\multirow{4}{*}{\rotatebox{90}{\makecell{\emph{boxes}}}} 
& CMax \cite{Gallego18cvpr} & 7.38 & 6.66 & 6.03 & 9.04 & 9.08 & 0.66\\
& AEMin \cite{Nunes20eccv} & 6.75 & 5.19 & 5.78 & 7.77 & 7.81 & 0.56\\
& EMin \cite{Nunes20eccv} & \textbf{6.55} & 4.40 & 5.00 & 7.00 & 7.06 & 0.51 \\ \addlinespace[0.02cm]
& Poisson Point-Proc. & 6.72 & \textbf{3.93} & \textbf{4.55} & \textbf{6.64} & \textbf{6.73} & \textbf{0.49}\\
\midrule 
\multirow{4}{*}{\rotatebox{90}{\makecell{\emph{poster}}}} 
& CMax \cite{Gallego18cvpr} & 13.45 & 9.87 & 5.56 & 13.39 & 13.45 & 0.74\\
& AEMin \cite{Nunes20eccv} & 12.57 & 7.89 & 5.63 & 12.35 & 12.36 & 0.68\\
& EMin \cite{Nunes20eccv} & 11.83 & 7.31 & 4.37 & 10.85 & 10.86 & 0.60 \\ \addlinespace[0.02cm]
& Poisson Point-Proc. & \textbf{11.78} & \textbf{6.33} & \textbf{3.67} & \textbf{10.30} & \textbf{10.37} & \textbf{0.57}\\ 
\midrule
\multirow{4}{*}{\rotatebox{90}{\makecell{\emph{dynamic}}}}
& CMax \cite{Gallego18cvpr} & 4.93 & 4.82 & 4.95 & 7.11 & 7.13 & 0.71\\
& AEMin\cite{Nunes20eccv} & 5.02 & 3.88 & 4.55 & 6.16 & 6.19 & 0.62\\
& EMin \cite{Nunes20eccv} & 4.78 & 3.72 & 3.73 & 5.33 & 5.39 & 0.54\\
& Poisson Point-Proc. & \textbf{4.42} & \textbf{3.61} & \textbf{3.49} & \textbf{5.15} & \textbf{5.19} & \textbf{0.52}\\
\midrule
\multirow{4}{*}{\rotatebox{90}{\makecell{\emph{shapes}}}}
& CMax \cite{Gallego18cvpr} & 31.19 & 26.83 & 38.98 & 55.86 & 55.87 & 3.94 \\
& AEMin \cite{Nunes20eccv} & 22.22 & 18.78 & 35.41 & 55.43 & 55.44 & 3.91 \\
& EMin \cite{Nunes20eccv} & 21.22 & 15.87 & 25.57 & 42.22 & 42.22 & 2.98 \\
& Poisson Point-Proc. & \textbf{20.73} & \textbf{13.95} & \textbf{17.69} & \textbf{25.88} & \textbf{25.89} & \textbf{1.83} \\
\bottomrule
\end{tabular}
\end{adjustbox}
\vspace{0.5ex}
\caption{
{\bf Angular velocity estimation}. 
Accuracy comparison on the rotation sequences from dataset~\cite{Mueggler17ijrr}.}
\label{table:main}
\end{table}

\paragraph{Evaluation metrics.}
The dataset~\cite{Mueggler17ijrr} does not come with an associated evaluation protocol. We therefore define an evaluation protocol in accordance to previous work for angular velocity estimation.
Typically algorithms for angular velocity estimation estimate a constant velocity $\omega$ over a fixed set of $N$ events. Let $t_\text{start}$ be the time stamp of the first event within the set of $N$ events and $t_\text{end}$ be the time stamp of the last event.
We compare the estimated velocity $\omega$ with the ground truth at time $t_\text{mid}=(t_\text{end}-t_\text{start})/2$.
Similar to~\cite{Nunes20eccv} we evaluate all methods using four different error measurements: angular velocity error $(e_{\omega_x}, e_{\omega_y}, e_{\omega_z})$ in degree/s, their standard deviation $\sigma_{e_\omega}$, the RMS-error in degree/s. The RMS error compared to the maximum excursions of ground truth is presented as a percentage (\%).

We also show results on linear velocity estimation of the camera. 
Since the depth of the scene is not provided~\cite{Mueggler17ijrr}, linear velocity with a single camera can only be estimated up to scale. 
In this case we compare the estimated and ground truth linear velocities by computing scale factor between them (via linear regression). 

\textbf{Ground truth}. 
The built-in gyroscope of the DAVIS' IMU~\cite{Delbruck14iscas} provides accurate measurements of the camera's orientation and therefore is used in this paper to evaluate the camera's angular velocity. 
As reported by~\cite{Mueggler17ijrr}, measurements of the IMU come with a temporal lag of $\approx2.4$ms, which we take into account in our evaluation pipeline.
On the other hand, the quality of the data produced by the DAVIS' IMU accelerometer does not match the high positional accuracy of the motion capture system. 
Hence we use latter for linear velocity assessment.


\subsection{Results}
\label{sec:experim:results}

\paragraph{Angular velocity estimation.}
We compare our method (maximization of likelihood~\eqref{eq:loss:polarity}) to the most recent work for angular velocity estimation~\cite{Gallego18cvpr,Nunes20eccv}. Gallego et al.~\cite{Gallego18cvpr, Gallego19cvpr} estimate motion by maximizing the contrast (e.g., variance) of an image of warped events (IWE). 
Nunes et al.~\cite{Nunes20eccv} estimate motion by minimization of an entropy (e.g. Tsallis') defined between pairs of events in the spatio-temporal volume. 
They provide an exact entropy calculation, which is expensive, and an approximate one, which is faster. 
We compare against both, in terms of accuracy and runtime.

Table \ref{table:main} shows the quantitative comparison of accuracy among all event-based angular velocity estimation methods on all four rotational motion sequences. 
Our Poisson point process method consistently outperforms the baseline methods. 
On poster\_rotation and boxes\_rotation our approach shows an improvement of about $5\%$ measured based on the root mean square error (RMS). On the other two sequences we show improvement of $4\%$ and $39\%$ compared to the next best performing method. The gain of $39\%$ in particular shows our superior performance for event recordings where the scene structure (brightness) and motion varies significantly. The shapes sequence contains much fewer events due to its ``simple'' scene structure than the other three sequences of this dataset.
Even during peak velocities of about $\pm 940$~deg/s, which corresponds to 2.5 full rotations per second, our approach robustly estimates the motion. 

Figure.~\ref{fig:results} shows qualitative results for all four approaches used for comparison. We show the ground truth aligned event image together with a detailed close-up view for each approach highlighting the alignment quality of events at object edges. Our approach consistently reconstructs very `sharp' object contours (see first three rows of Fig.~\ref{fig:results}). The shapes sequence (depicted in the last column of Fig.~\ref{fig:results}) comes with quite different image characteristics. Due to its rather simple structure, this event sequence
comprises roughly 20\% of the amount of events that are usually acquired during the same time. Therefore a fixed number of events that is consistently used over all four video sequences results in much larger time intervals containing highly varied motion. In these cases none of the algorithms is able to align the events accurately assuming constant velocity during a fixed number of events. 


\global\long\def\figWidth{0.11\linewidth}
\global\long\def\figWidthFirst{0.145\linewidth}
\global\long\def\figWidthLast{0.128\linewidth}
\begin{figure*}[t]
	\centering
    {\small
    \setlength{\tabcolsep}{2pt}
	\begin{tabular}{
	>{\centering\arraybackslash}m{0.3cm} 
	>{\centering\arraybackslash}m{\figWidthFirst}
	>{\centering\arraybackslash}m{\figWidthFirst} 
	>{\centering\arraybackslash}m{\figWidth}
	>{\centering\arraybackslash}m{\figWidth} 
	>{\centering\arraybackslash}m{\figWidth}
	>{\centering\arraybackslash}m{\figWidth} 
	>{\centering\arraybackslash}m{\figWidthLast}}
	& Grayscale frame & Aligned using IMU & Unaligned & CMax & AEMin & EMin & Ours \\\addlinespace[0.5ex]
	\rotatebox{90}{\makecell{\emph{dynamic}}}
	&\includegraphics[width=\linewidth]{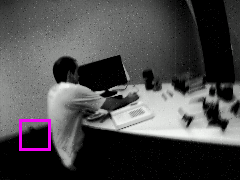}
    &\includegraphics[width=\linewidth]{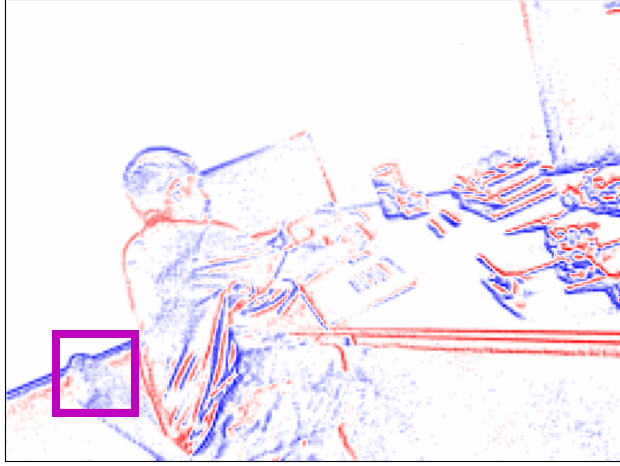}
    &\includegraphics[width=\linewidth]{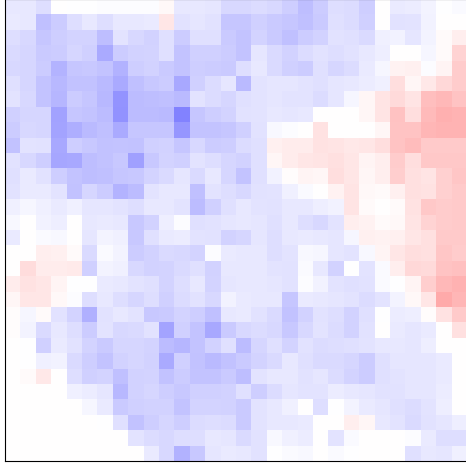}
    &\includegraphics[width=\linewidth]{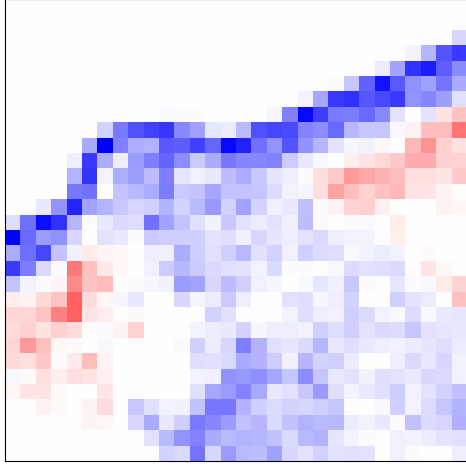}
    &\includegraphics[width=\linewidth]{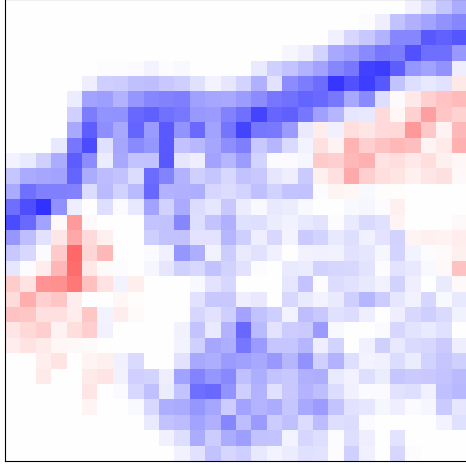}
    &\includegraphics[width=\linewidth]{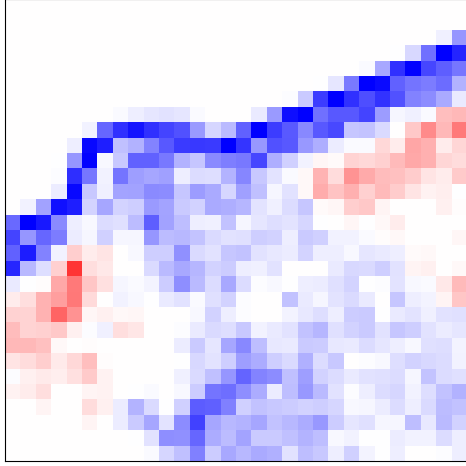}
    &\includegraphics[width=\linewidth]{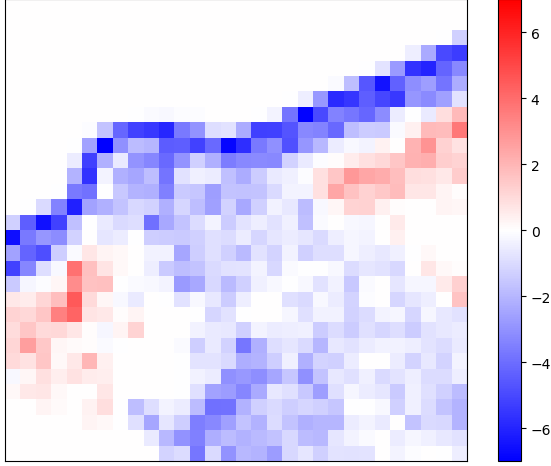}
	\\[-0.5ex]
	\rotatebox{90}{\makecell{\emph{boxes}}}
	&\includegraphics[width=\linewidth]{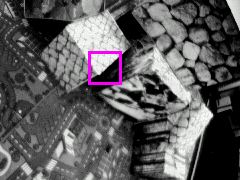}
    &\includegraphics[width=\linewidth]{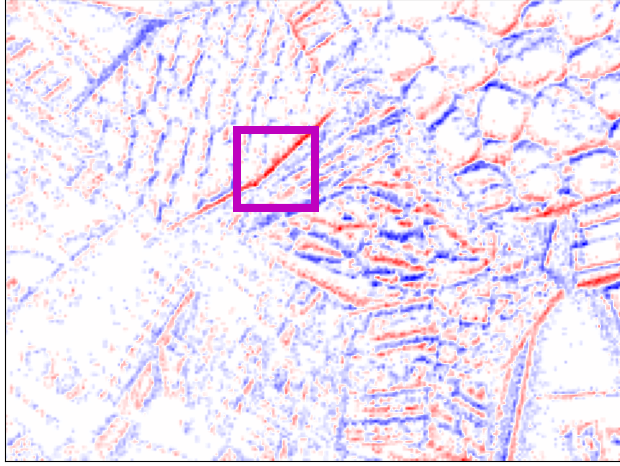}
    &\includegraphics[width=\linewidth]{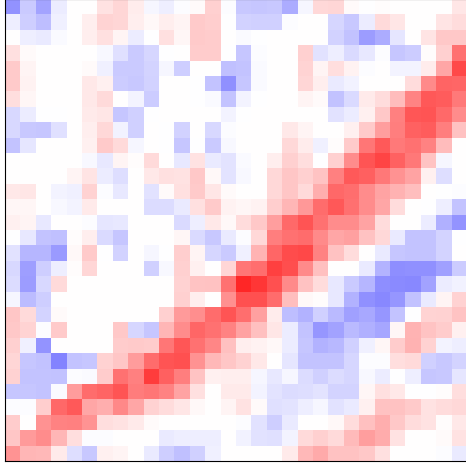}
    &\includegraphics[width=\linewidth]{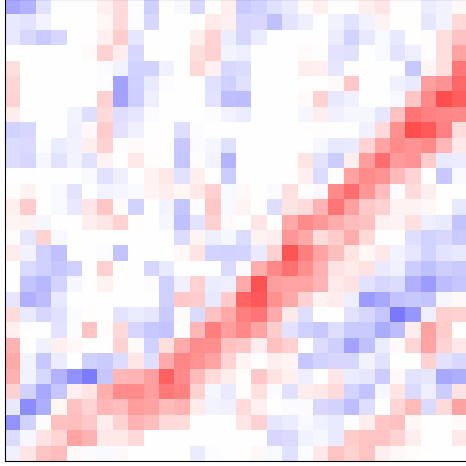}
    &\includegraphics[width=\linewidth]{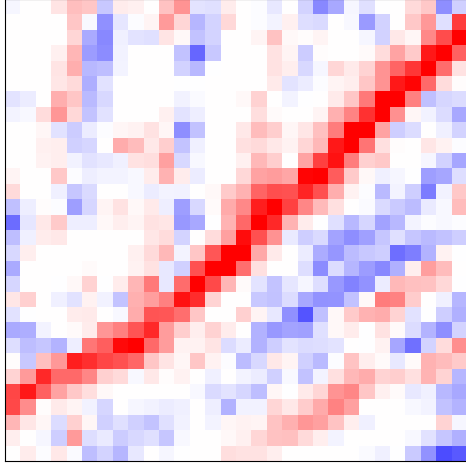}
    &\includegraphics[width=\linewidth]{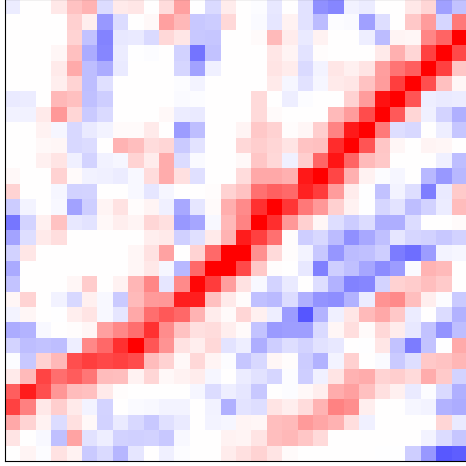}
    &\includegraphics[width=\linewidth]{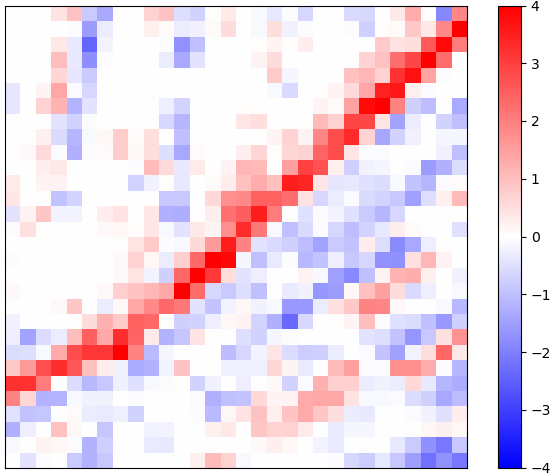}
    \\[-0.5ex]
    \rotatebox{90}{\makecell{\emph{poster}}}
    &\includegraphics[width=\linewidth]{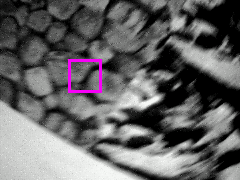}
	&\includegraphics[width=\linewidth]{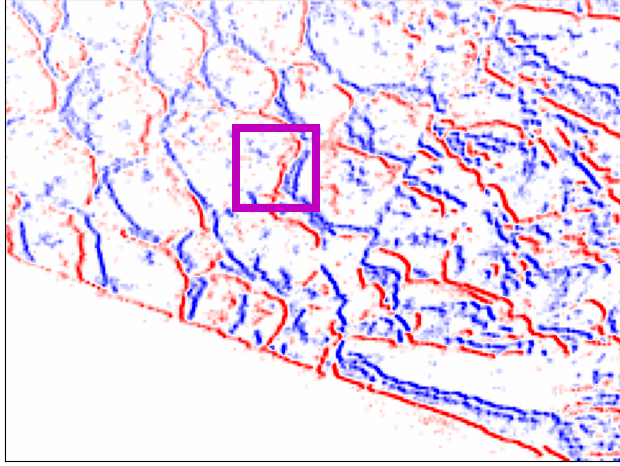}
    &\includegraphics[width=\linewidth]{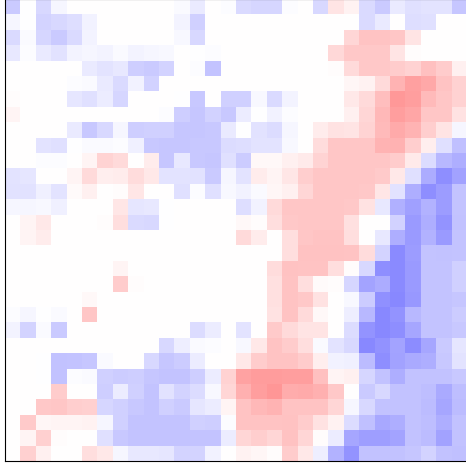}
    &\includegraphics[width=\linewidth]{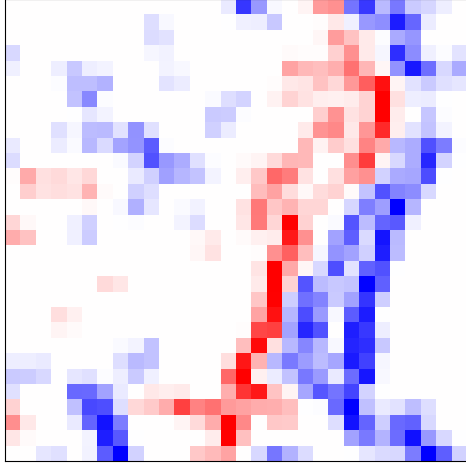}
    &\includegraphics[width=\linewidth]{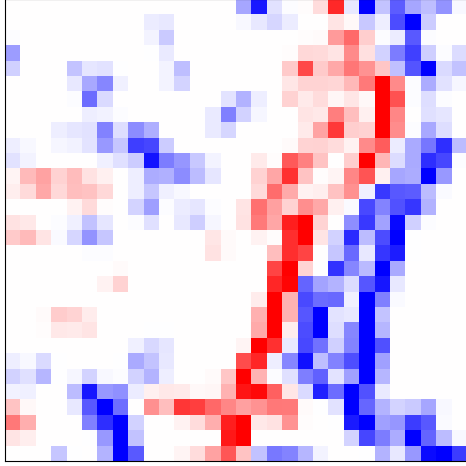}
    &\includegraphics[width=\linewidth]{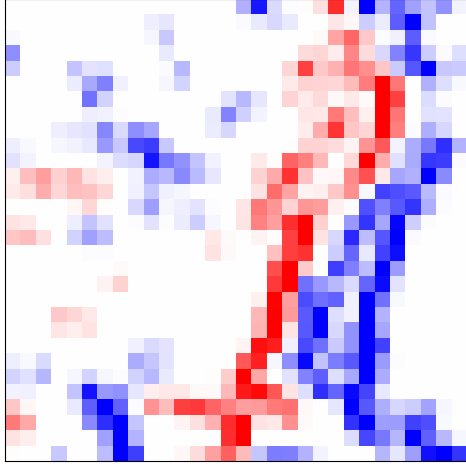}
    &\includegraphics[width=\linewidth]{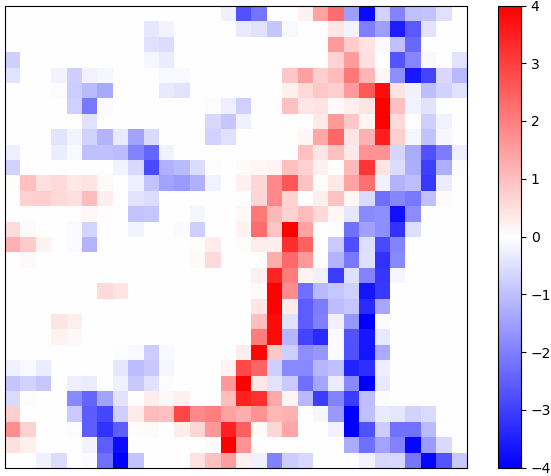}
    \\[-0.5ex]
    \rotatebox{90}{\makecell{\emph{shapes}}}
    &\includegraphics[width=\linewidth]{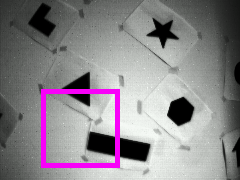}
    &\includegraphics[width=\linewidth]{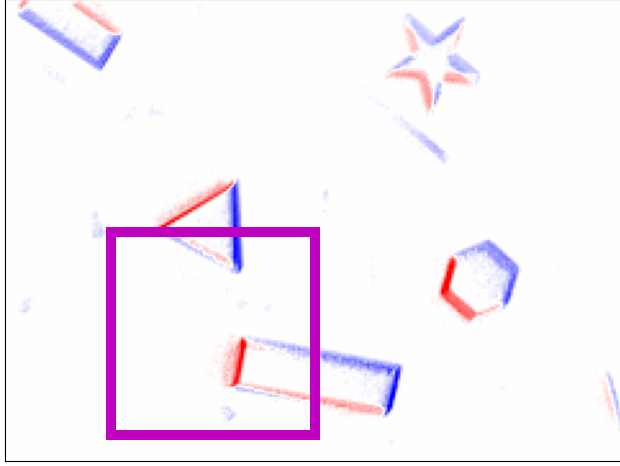}
    &\includegraphics[width=\linewidth]{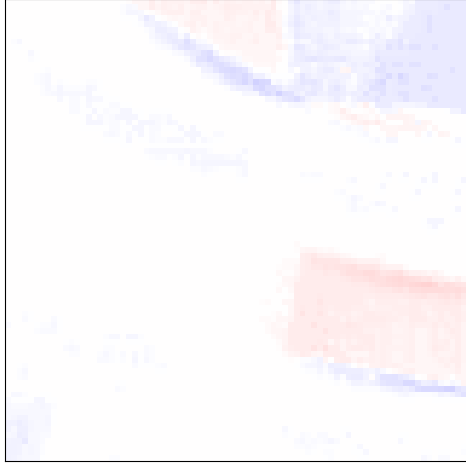}
    &\includegraphics[width=\linewidth]{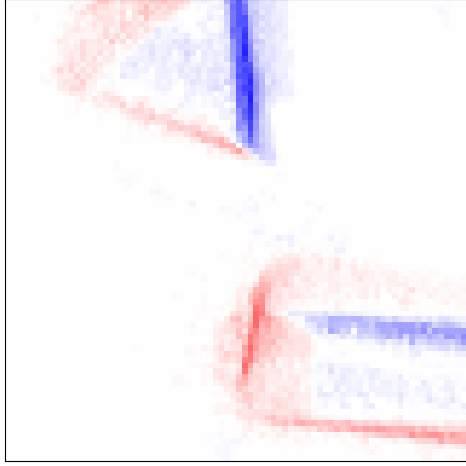}
    &\includegraphics[width=\linewidth]{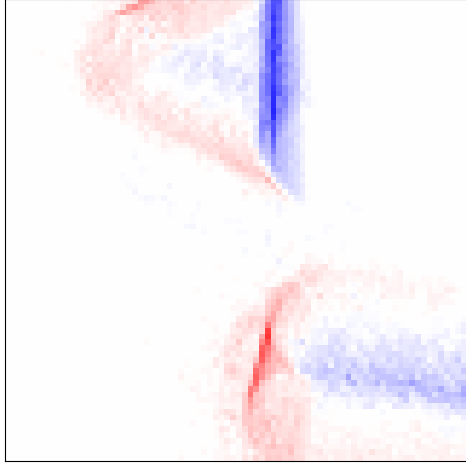}
	&\includegraphics[width=\linewidth]{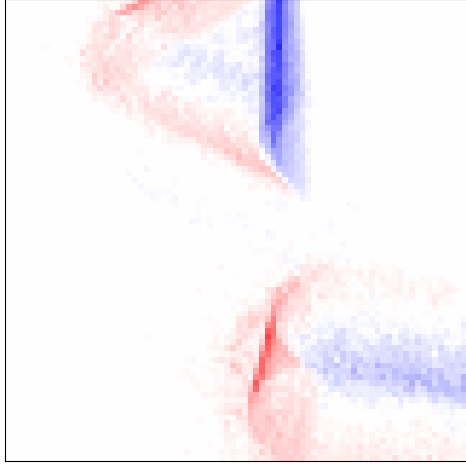}
	&\includegraphics[width=\linewidth]{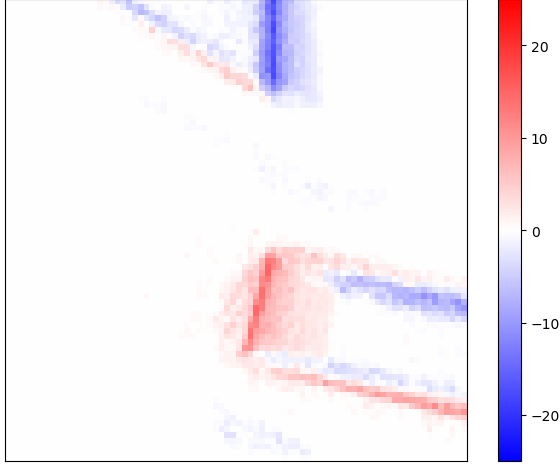}
	\\
	\end{tabular}
	}
\caption{\textbf{Qualitative results.} We show qualitative results for each video sequence with dominating rotational camera motion. 
From top to bottom: dynamic\_rotation, boxes\_rotation, poster\_rotation and shapes\_rotation. 
From left to right: Grayscale frame, aligned event image using ground truth from IMU, unaligned events, CMax, AEMin, EMin, Ours. 
Note that the \emph{shapes} sequence comprises a sparser scene texture, thus a batch of a fixed number of events spans over a larger time interval with more variation in camera motion than the other three, more textured, sequences. In this sequence, accurate alignment is not possible for any of the methods under the constant velocity assumption. 
The affine model of angular velocity that we propose in Section~\ref{sec:experim:affineomega} mitigates this issue.
}
\label{fig:results}
\end{figure*}

\textbf{Linear velocity estimation.}
Since our approach is flexible to the type of spatial transformation considered, 
we also assess its performance on the estimation of translational camera motion, e.g., linear velocity.
Table~\ref{table:main-linear-velocity} summarizes the results of event-based linear velocity estimation using also non-overlapping packets of 30k events.
For this task, we use the same textured scenes in~\cite{Mueggler17ijrr}, but the set of sequences with translational motion.
A challenge for all methods evaluated here is to avoid all events warping to a single pixel (undesired minima of the alignment measures), which can happen for large $Z$-motions.

Additional visual results for velocity estimation are provided in supplementary material.

\begin{table}
\centering
\small
\begin{adjustbox}{max width=\linewidth}
\setlength{\tabcolsep}{3pt}
\begin{tabular}
{l*{6}{c}}
\toprule 
\bf{Method} & $e_{vx}$ & $e_{vy}$ & $e_{vz}$ & $\sigma_{ev}$ & RMS & RMS\%\\
\midrule 
CMax \cite{Gallego18cvpr} & 0.21 & 0.26 & 0.41 & 0.42 & 0.43 & 7.83\\
AEMin \cite{Nunes20eccv} & 0.21 & 0.25 & 0.42 & 0.44 & 0.45 & 8.36\\
EMin \cite{Nunes20eccv} & 0.35 & 0.43 & 0.46 & 0.63 & 0.64 & 11.80\\ \addlinespace[0.02cm]
Poisson Point-Proc. & \textbf{0.17} & \textbf{0.22} & \textbf{0.38} & \textbf{0.38} & \textbf{0.38} & \textbf{6.93}\\
\bottomrule
\end{tabular}
\end{adjustbox}
\vspace{0.5ex}
\caption{{\bf Linear velocity estimation.} 
Accuracy comparison on four translational sequences from dataset~\cite{Mueggler17ijrr}. Average results.
}
\label{table:main-linear-velocity}
\end{table}

\begin{figure}[b]
    \centering
    \includegraphics[width=0.8\columnwidth]{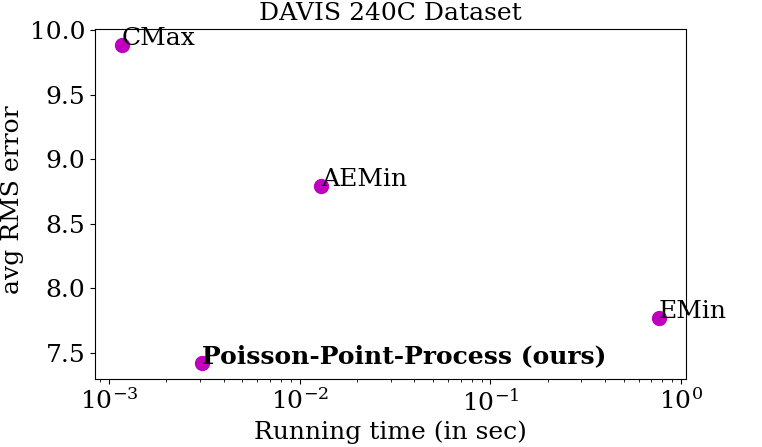}
    \caption{\textbf{Runtime vs.~accuracy comparison.} Time for one loss function evaluation given a fixed set of 30k events versus accuracy measured in terms of average RMS-error across boxes, poster and dynamic. Time plotted on log-scale.}
    \label{fig:runTimeVSAccuracy}
\end{figure}

\paragraph{Runtime and time complexity analysis.} 
We measure the time that it takes to compute the alignment (loss) function given a set of 30k events. 
For comparability we re-implement our loss function in C++. 
The runtime was measured using an 8-core CPU with 16 threads and clock speed of 3.9 GHz.
The runtime of all four tested methods is compared in Fig.~\ref{fig:runTimeVSAccuracy}, and plotted against accuracy. Our here proposed approach achieves highest average accuracy for angular velocity estimation and is among the fastest (3.1ms for one loss function evaluation). The contrast maximization approach is the fastest approach taking just 1.2ms per function evaluation, however comes with significantly lower performance in terms of the RMS-measure.

Additionally, the complexity analysis in Table~\ref{table:complexity} explains the slow computation time of both entropy minimization methods (EMin, AEMin). 
The complexity of our approach as well as for contrast maximization is linear with the number of events $N_e$. 
The complexity of Emin~\cite{Nunes20eccv} is quadratic with the number of events since it requires the evaluation of costs due to all pairs of events. 
The faster, approximate version of EMin only considers costs due to events within a certain distance defined by a Kernel, of size $\kappa^d$, thus reducing the complexity from $N^2_e$ to $N_e\kappa^d$, where $\kappa\ll N_e$.

\begin{table}[t]
\centering
\small
\begin{tabular}
{lc}
\toprule 
\bf{Method} & \bf{Time complexity} \\
\midrule 
CMax \cite{Gallego18cvpr} & $\cO(N_e)$ \\
AEMin \cite{Nunes20eccv}  &  $\cO(N_e \kappa^d)$\\
EMin \cite{Nunes20eccv}  & $\cO(N^2_e)$ \\ \addlinespace[0.02cm] 
Poisson Point-Process (ours) & $\cO(N_e)$ \\
\bottomrule
\end{tabular}
\vspace{1ex}
\caption{{\bf Time complexity} of each algorithm as a function of the number of input events $N_e$ and kernel size $\kappa^d$.}
\label{table:complexity}
\end{table}

\subsection{Ablation study}
\paragraph{Poisson rate parameter $\lambda$ - the expected rate of events.}\label{subsec:lambdaprior}
In Section~\ref{sec:optim-problem} we have described our optimization problem as a maximum likelihood procedure: 
the likelihood of aligned data modeled as a Poisson point process $\text{Pois}(\lambda_\x)$ is higher than the likelihood of unaligned data under our model. 
Computing the likelihood of events requires knowledge of the rate parameter $\lambda$.
Here, we discuss two options to deal with this unknown parameter: 
($i$) marginalizing it out,
($ii$) using its per-pixel ML-estimate. 

\begin{figure}[t]
    \centering
    \includegraphics[width=0.85\columnwidth]{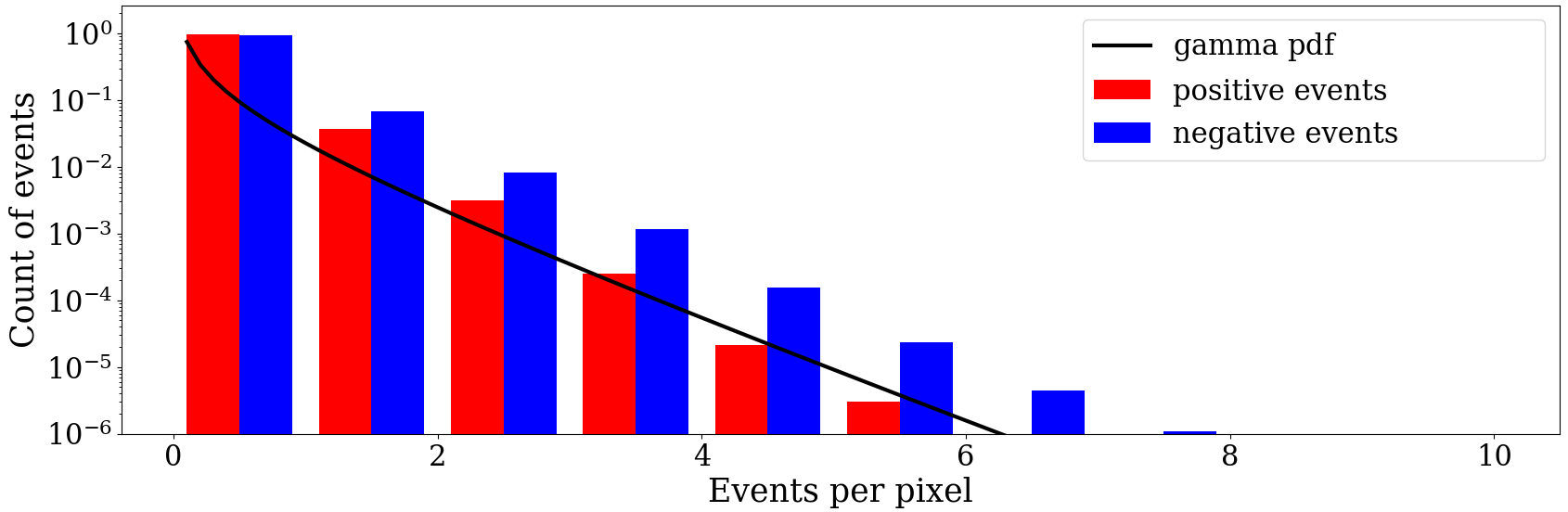}
    \caption{\textbf{Prior distribution over $\lambda$.} Histogram of expected event counts per pixel $\lambda_{\mathbf{x}}$ during a time interval $\Delta T$ (events with positive polarity in red and negative in blue). 
    PDF of the Gamma distribution shown overlaid.
    }
    \label{fig:histogram}
\end{figure}

\emph{Marginalizing out $\lambda$.}
Integrating over $\lambda$ leads to a negative binomial distribution (Eq.~\ref{eq:loss:polarity}) - which is also often described as a gamma-Poisson mixture with a gamma distribution as the mixing distribution. 
We derive our prior distribution $\lambda \sim \text{Gamma}(r, q^{-1}(1-q))$ from observed (unaligned) event data. In particular, the data is the expected counts of events per pixel during a time interval $\Delta T$. 
Both parameters $r$ and $q$ defining the Gamma distribution are obtained via maximum likelihood estimation. Fig.~\ref{fig:histogram} shows the histogram of expected event counts per pixel for a set of 30000 events. The best fitting Gamma distribution with parameters $r=0.1$ and $q=0.39$ is shown overlaid.

\emph{Per-pixel ML-estimate, $\lambda_\x$}. Given a set of events at a particular pixel location $\x$, the ML-estimate for the rate parameter $\lambda_\x$ is simply the \emph{count} of events at that location (since we just have one observation sample).
This approach might have the advantage of capturing the scene structure, where a point in the world triggers events at different rates. However due to the relatively small sample size this approach is less robust than integrating over the unknown variable.

Overall both approaches perform well, but marginalizing the unknown variable out seems to be more robust on average. Using an ML-estimate for $\lambda$ leads to an average RMS-error of 12.1 deg/s, marginalisation improves slightly and reaches an RMS-error of 12.0 deg/s.

\begin{figure}[h]
\begin{center}
    \subfigure[Grayscale frame]
    {
        \includegraphics[height=0.22\linewidth]{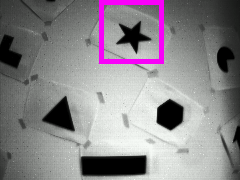}
    }
    \subfigure[Alignment (IMU) - \emph{typical eval. freq.}]
    {
        \includegraphics[height=0.22\linewidth]{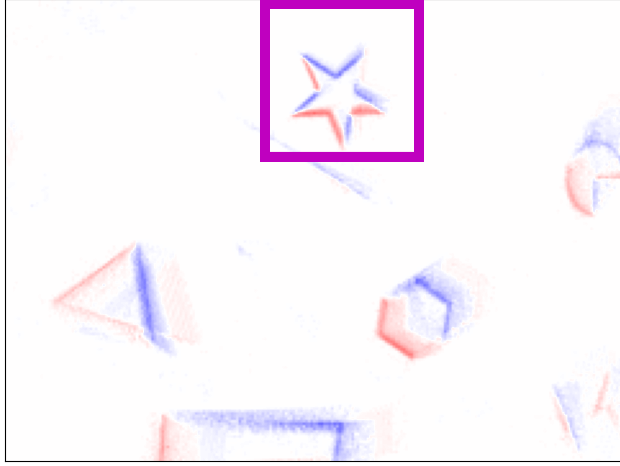}
        \label{fig:events-aligned-gt-costVel}
    }
    \subfigure[Alignment (IMU) - \emph{ground truth freq.}]
    {
        \includegraphics[height=0.22\linewidth]{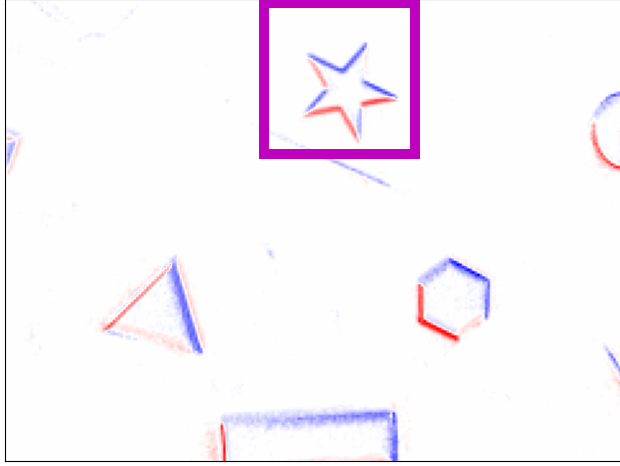}
    }
    \\
    \subfigure[Unaligned]
    {
        \includegraphics[height=0.285\linewidth]{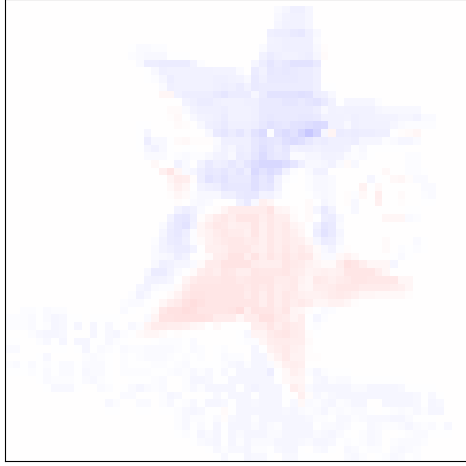}
    }
    \subfigure[Alignment (our) - \emph{const. velocity}]
    {
        \includegraphics[height=0.285\linewidth]{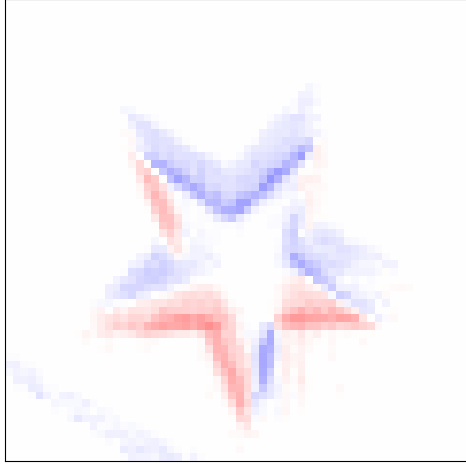}
    }
    \subfigure[Alignment (our) - \emph{affine velocity model}]
    {
        \includegraphics[height=0.285\linewidth]{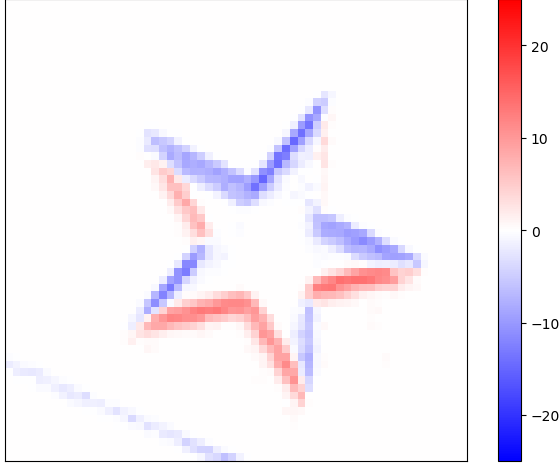}
        \label{fig:events-aligned-slope}
    }
   \caption{\textbf{Affine velocity model}. 
   Quality of event alignment for constant velocity estimates within $\Delta T$ compared to \emph{smooth} velocity model estimates from our affine velocity model.}
   \label{fig:slope}
\end{center}
\end{figure}

\paragraph{Affine model of angular velocity.}
\label{sec:experim:affineomega}
Event alignment of a fixed batch of events (e.g., 30k) is typically done via assuming a constant velocity during the time span of the events. 
However, such a time span is a variable that depends on the amount of texture in the scene. 
As the last row of Fig.~\ref{fig:results} shows, 30k events deem too many for low-textured scenes (shapes). 
A possible fix to this issue is to use an adaptive number of events, depending on texture~\cite{Liu18bmvc}. 
However this makes comparisons more difficult to interpret. 
We develop a different solution: using a more expressive motion model.
Fig.~\ref{fig:slope} shows that for a large interval $\Delta T$ a high quality alignment can only be achieved with more complex (but smooth) velocity estimates, such as the ground truth signal $\omega(t)$.
Since alignment with constant velocity $\omega(t)\approx\omega_0\;\forall t\in[0,\Delta T]$ is not enough, we propose a higher order model $\omega(t)\approx\omega_0+a t$ (affine), 
thus estimating $(\omega_0,a)\in\mathbb{R}^6$.
This improves event alignment (Fig.~\ref{fig:events-aligned-slope}).
A typically evaluation strategy is to compare the estimated \emph{constant} velocity with its closest ground truth. This allows for evaluating average angular velocity over a fixed time interval, however leads to inaccurate event alignment as can be seen in Figure~\ref{fig:events-aligned-gt-costVel}. To mitigate this issue evaluating angular velocity using the high frequency (1kHz) of the ground truth provided by~\cite{Mueggler17ijrr} is required.

\section{Conclusion}
Inspired by congealing~\cite{LearnedMiller06pami}, this paper has introduced a new probabilistic approach for event alignment.
In particular we model the aligned events as independent, \emph{per-pixel Poisson point processes}, or a spatio-temporal Poisson point process. 
Based on this idea, we derive a likelihood function for a set of observed (unaligned) events and maximize it to estimate the camera motion that best explains the events.
This method leads to new state-of-the-art results for angular velocity estimation, with only 0.5\% relative RMS error with respect to the velocity excursion.
Our event alignment method is not specific of rotational motion, 
as we have demonstrated how it can be applied to other types of motion (e.g., translational). 
This opens the door to utilize our method for solving related event-alignment problems, such as motion segmentation~\cite{Stoffregen19iccv} and feature tracking~\cite{Seok20wacv}, which in turn enable higher level scene understanding.

{\small
\bibliographystyle{ieee_fullname}
\bibliography{paper}
}

\appendix
\clearpage
\section{Supplementary Material}
We show additional visual results indicating the high quality of our proposed method and provide a more theoretical background of our model justifying the likelihood function of observed event data. The observed event data is defined through the mapping $(\x, t) \mapsto (R_\omega^t  \x, t)$, where $R_\omega^t$ defines the mapping that is applied to event data. This mapping can be formally justified via the Poisson mapping theorem, which is discussed here.

\section{Visual results for velocity estimation}
\subsection{Angular Velocity}
\begin{figure*}
\centering
 \subfigure[Angular velocity around x-axis: full sequence (60sec)]{
     \includegraphics[width=\textwidth]{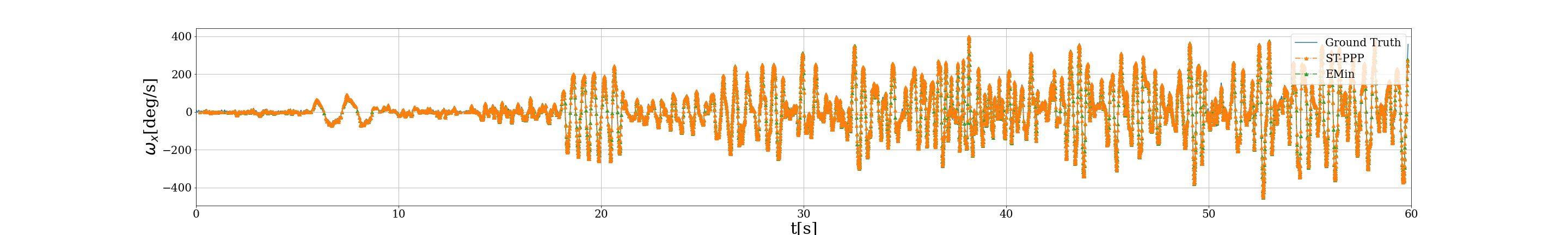}
     \label{fig:angular-vel-a} }
 \subfigure[Zoomed-in plots of corresponding bounded regions]{
     \includegraphics[width=\textwidth]{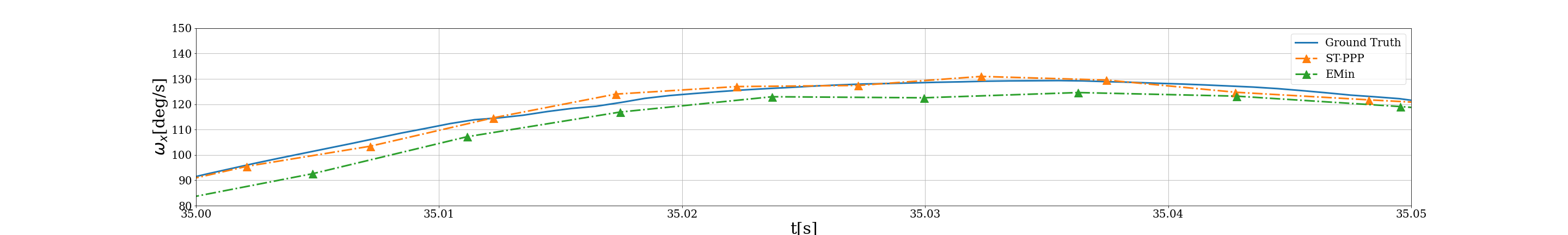}
     \label{fig:angular-vel-b} }
 \subfigure[Angular velocity around y-axis: full sequence (60sec)]{
     \includegraphics[width=\textwidth]{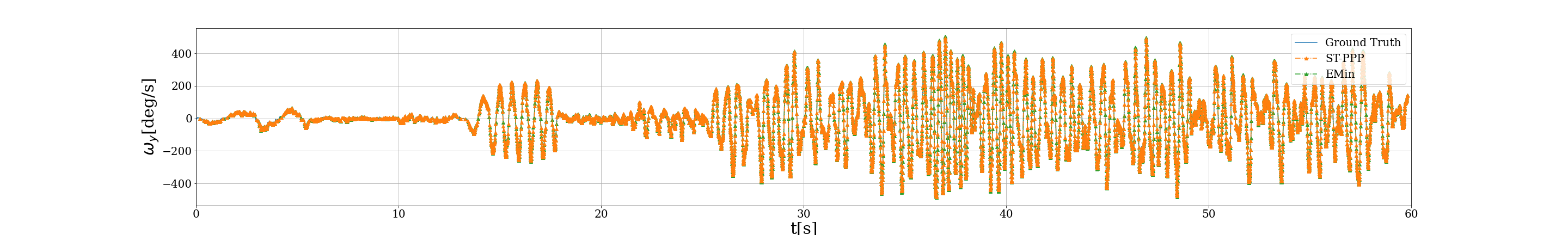}
     \label{fig:angular-vel-c} }
\subfigure[Zoomed-in plots of corresponding bounded regions]{
     \includegraphics[width=\textwidth]{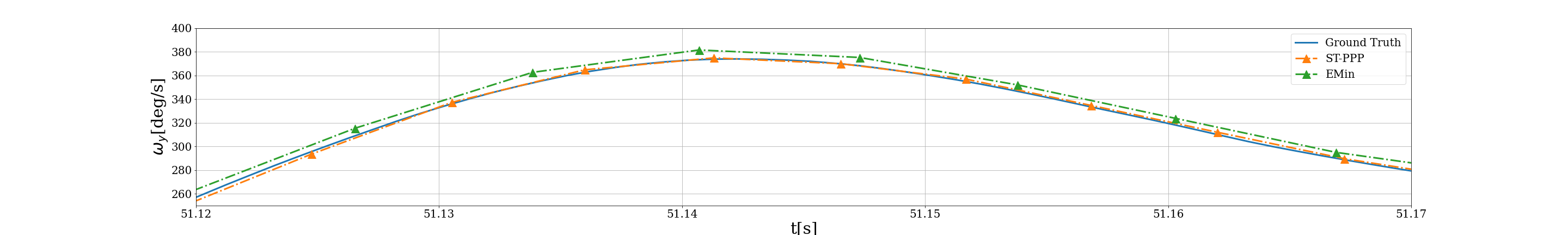}
     \label{fig:angular-vel-d} }
 \subfigure[Angular velocity around z-axis: full sequence (60sec)]{
     \includegraphics[width=\textwidth]{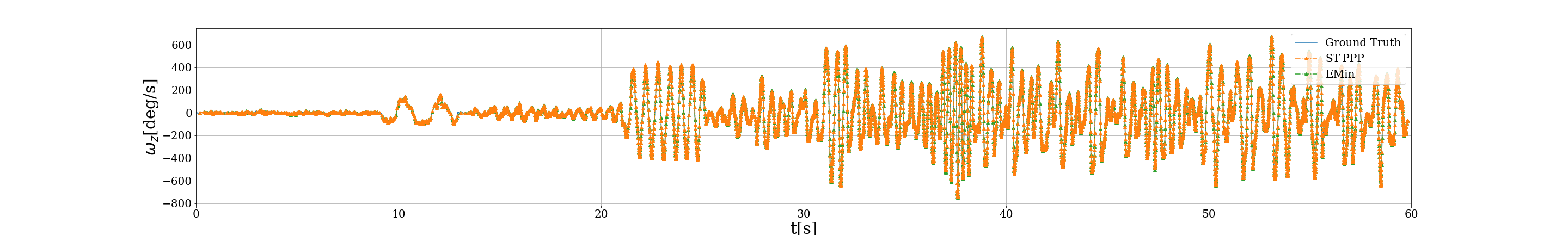}
     \label{fig:angular-vel-e} }
 \subfigure[Zoomed-in plots of corresponding bounded regions]{
     \includegraphics[width=\textwidth]{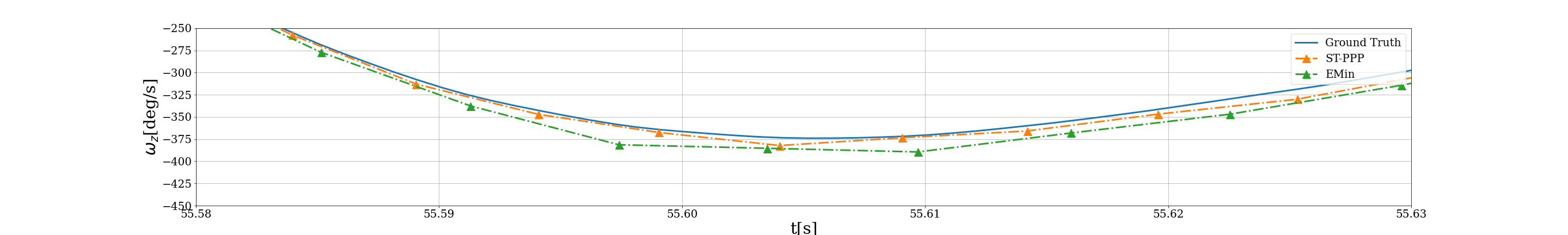}
     \label{fig:angular-vel-f} }
\caption{\textbf{Angular velocity estimates} measured in deg/sec plotted versus ground truth (IMU). 
Example \emph{boxes\_rotation}. Comparison to the next best performing method (EMin).
}
\label{fig:results-angular-velocity}
\end{figure*}

In Figure~\ref{fig:results-angular-velocity} we show the accuracy of our estimated velocity compared to the next best performing method (EMin~\cite{Nunes20eccv}) and ground truth (IMU). Taking the sequence boxes\_rotation as an example we show the estimated angular velocities over the entire sequence (60sec) as well as over a shorter duration (0.05sec) for each axis of rotation.
Even during peak velocities with around $380 \text{ deg}/\text{sec}$ estimates obtained by our method are robust and outperform previous results from~\cite{Nunes20eccv} (Figure~\ref{fig:angular-vel-d}).

\begin{figure}
    \centering
    \includegraphics[width=0.75\columnwidth]{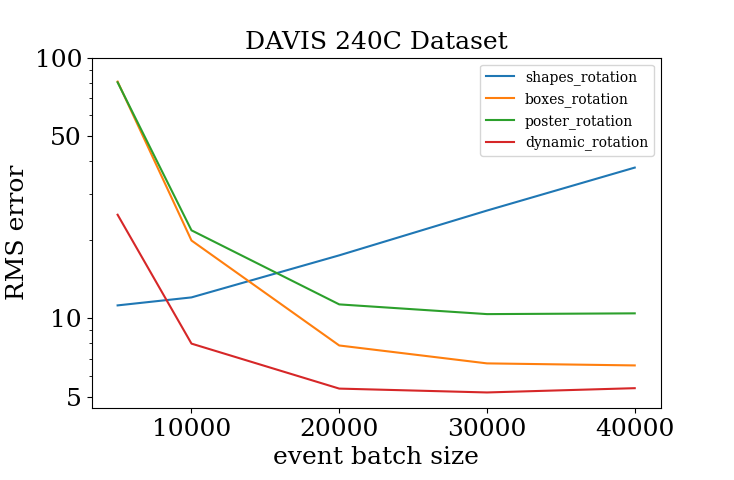}
    \caption{\textbf{Event batch size vs.~accuracy.} Accuracy measured in terms of RMS error (deg/sec) for different event batch sizes to process the four rotational sequences from dataset~\cite{Mueggler17ijrr}.}
    \label{fig:batchSize}
\end{figure}

\paragraph{Constant velocity assumption within a fixed event batch.}
Event alignment of a fixed batch of events (e.g., 30k) is typically done via assuming a constant velocity during the time span of the events. 
However, such a time span is a variable that depends on the amount of texture in the scene and motion of the camera. 
A possible fix to this issue is to use an adaptive number of events, depending on texture~\cite{Liu18bmvc}. However this makes comparisons more difficult to interpret. 

In Figure~\ref{fig:batchSize} we show for each sequence the accuracy reached for a specific batch size of events. We vary the batch size between 5k and 40k events. While the three sequences boxes, poster and dynamic show relatively realistic scenes, the shapes sequence shows just a few black shapes posted onto a white background. Due to this rather simple texture much fewer events are generated within the same time interval. Conversely, a fixed number of 30k events spans over a much larger time interval, with possibly high variation in motion. If the batch size of events is too large, the constant velocity assumption leads to a significant drop in performance. As one can see in Figure~\ref{fig:batchSize} velocity estimations based on fewer events are significantly more suitable for the shape\_sequence.

\subsection{Linear Velocity}
\begin{figure*}
\centering
 \subfigure[Linear velocity around x-axis: full sequence (60sec)]{
     \includegraphics[width=\textwidth]{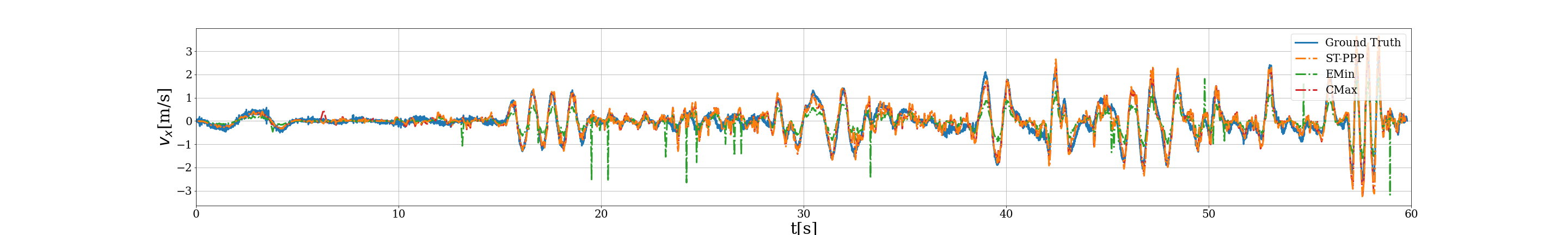}
     \label{fig:linear-vel-a} }
 \subfigure[Zoomed-in plots of corresponding bounded regions]{
     \includegraphics[width=\textwidth]{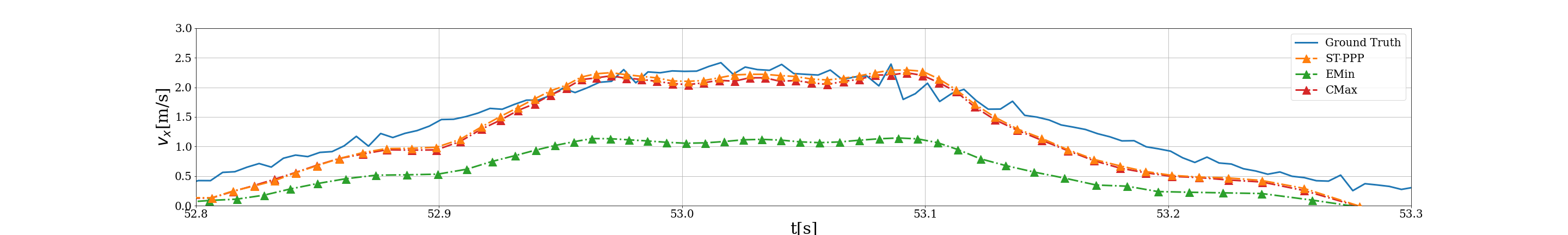}
     \label{fig:linear-vel-b} }
 \subfigure[Linear velocity around y-axis: full sequence (60sec)]{
     \includegraphics[width=\textwidth]{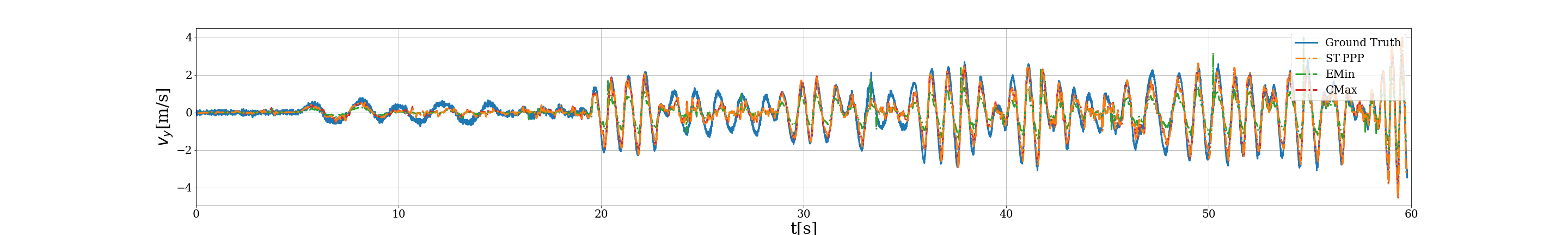}
     \label{fig:linear-vel-c} }
\subfigure[Zoomed-in plots of corresponding bounded regions]{
     \includegraphics[width=\textwidth]{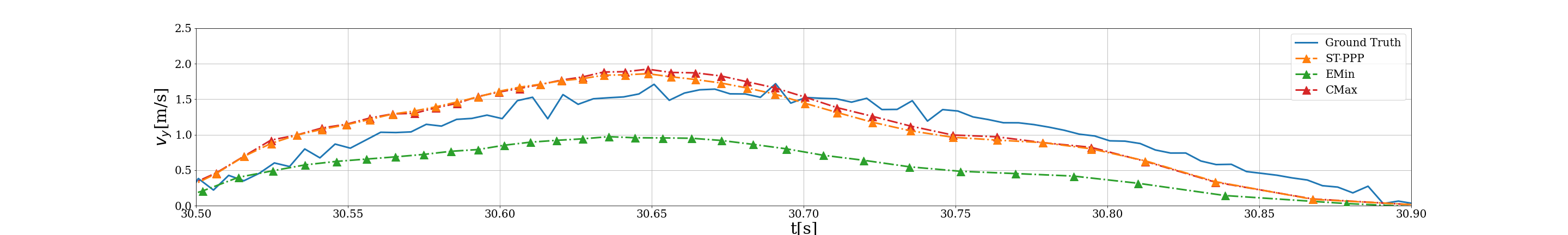}
     \label{fig:linear-vel-d} }
 \subfigure[Linear velocity around z-axis: full sequence (60sec)]{
     \includegraphics[width=\textwidth]{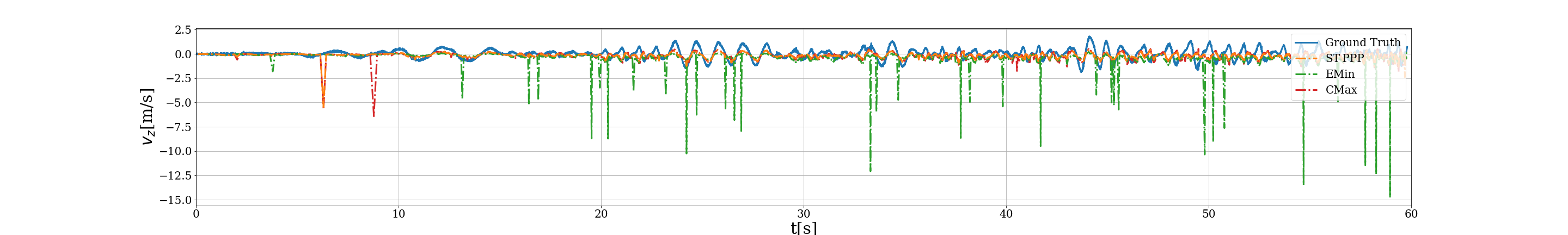}
     \label{fig:linear-vel-e} }
 \subfigure[Zoomed-in plots of corresponding bounded regions]{
     \includegraphics[width=\textwidth]{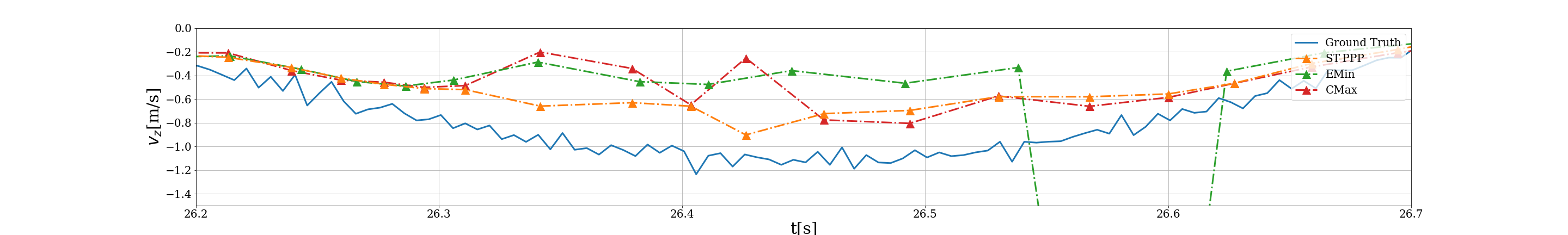}
     \label{fig:linear-vel-f} }
\caption{\textbf{Linear velocity estimates} measured in m/s plotted versus ground truth from motion capture system. 
Example sequence \emph{boxes\_translation}. 
Comparison to EMin~\cite{Nunes20eccv}, CMax~\cite{Gallego18cvpr} and ground truth.
}
\label{fig:results-linear-velocity}
\end{figure*}

Figure~\ref{fig:results-linear-velocity} shows the accuracy of EMin~\cite{Nunes20eccv}, CMax~\cite{Gallego18cvpr} and ours compared to ground truth for the boxes\_translation sequence. 
Ground truth is taken from a motion capture system. 
EMin trying to minimize the pairwaise entropy among \textit{all event-event pairs} suffers a lot from global minima which are reached for large $Z$-motions. 
In this case all events are mapped onto a single point, which is the focus of expansion of the camera. Besides being more robust to outliers it is visible that our new proposed method is also more accurate (see Figure~\ref{fig:linear-vel-b}, \ref{fig:linear-vel-d} and~\ref{fig:linear-vel-f}).


\section{Model - Additional Theoretical Background}

In this section, we elaborate on the definition of the likelihood $p_\cO(\cO | \omega) = p_\cA(R_\omega(\cA))$ as an instance of the Poisson mapping theorem. 
We use $q_\cA$ and $q_\cO$ to represent the density of point sets under Poisson processes, to distinguish from the notation $p_\cA$ and $p_\cO$ used in the main text for the probability of the resultant event counts.

\subsection{Density of aligned events $\cA$}
Recall that the aligned events $\cA$ are distributed according to a Poisson process on $\X \times [0, \Delta T]$ with intensity function $\lambda(\x, t) \doteq \Delta T^{-1} \lambda_\x$; the factor $\Delta T^{-1}$ adjusts for the time interval so that $k_\x \sim \text{Pois}(\lambda_\x)$, where $k_\x$ is the number of events observed at pixel $\x$ over the interval $[0, \Delta T]$.

The density of the point set $\cA = [(a_1^\x, a_1^t), \ldots, (a_N^\x, a_N^t)]$ is~\cite{Streit2010}
\begin{align}
q_\cA(\cA) &= \exp\Big(-\sum_{\x \in \X} \int_0^{\Delta T} \lambda(\x, t) dt \Big) \prod_{i=1}^N \lambda(a_i^\x, a_i^t) \\
& = \exp\Big(-\sum_{\x \in \X}\lambda_\x\Big)  \prod_{i=1}^N \Delta T^{-1} \lambda_{a_i^\x} \\
&= \Delta T^{-N} \prod_{\x \in \X} \lambda_\x^{k_\x}\exp(-\lambda_\x).
\end{align}
In the second line, we used $\int_0^{\Delta T} \lambda(\x, t) dt = \int_0^{\Delta T} \Delta T^{-1} \lambda_\x dt = \lambda_\x$. In the third line, we grouped events with $a_i^\x = \x$. These simplifications are possible because space is discrete and the intensity function is homogeneous with respect to time.

Observe that the probability $p_\cA(\cA)$ of the pixel counts as defined in the main text is related to $q_\cA(\cA)$ by
\begin{equation}
p_\cA(\cA) = \frac{\Delta T^N}{\prod_\x k_\x !} q_\cA(\cA) = \Big(\prod_\x \frac{\Delta T^{k_\x}}{k_\x!}\Big) q_\cA(\cA).
\end{equation}
The extra factor of $\prod_\x \frac{\Delta T^{k_\x}}{k_\x!}$ comes from integrating over all possible ordered sets of time indices $t_1, \ldots t_{k_\x} \in [0, \Delta T] $ for the $k_\x$ points for each pixel $\x$, and then dividing by $k_\x!$ to switch from an ordered tuple to an unordered set.

\subsection{Density of observed events $\cO$}

The Poisson mapping theorem describes what happens when the points of a Poisson process are mapped by a deterministic mapping: the result is a new Poisson process with modified intensity function. 
Let $f_t$ be the ground-truth mapping from reference coordinates to camera coordinates at time $t$. 
Assume for now that $f_t$ is a bijection on $\X$ for all $t$, as is the case for rotations. We discuss relaxations of this assumption below. Let $S$ be the joint mapping on space and time that sends $(\x, t)$ to $(f_t(\x), t)$, so the $i$th observed event is obtained from the $i$th aligned event as $o_i = S(a_i)$. By the Poisson mapping theorem, the observed point set $\cO = S(\cA) \doteq [S(a_1), \ldots, S(a_n)]$ is distributed according to a Poisson process with intensity function
\begin{align}
\lambda'(\x, t) &= \lambda\big( S^{-1}(\x, t)\big) = \lambda\big(f_t^{-1}(\x), t\big)\\
&= \Delta T^{-1} \lambda_{f_t^{-1}(\x)}.
\end{align}
In more general settings, a Jacobian term is required to adjust for changes of volume. In our case, it is not needed because the spatial coordinate is discrete, and the time coordinate mapping is the identity, which has unit Jacobian.

The density of the mapped point set $\cO$ is therefore
\begin{align}
q_\cO(\cO)
&= \exp\Big( -\sum_{\X} \int_0^{\Delta T} \lambda'(\x, t) dt\Big) \prod_{i=1}^N \lambda'(o_i^\x, o_i^t) \\
&= \exp\big(\sum_{\x \in \X} \lambda_\x \big) \prod_{i=1}^N \Delta T^{-1} \lambda_{f_t^{-1}(o_i^\x)} \\
&= q_\cA\big(S^{-1}(\cO)\big)
\end{align}
In the second line, we used $\sum_{\x \in \X} \lambda_{f_t^{-1}(\x)} = \sum_{\x \in \X} \lambda_\x$, which follows because $f_t$ is a bijection. 

By aggregating to counts in the same manner described above, we obtain the result used in the main text:
\begin{equation}
p_\cO(\cO) = p_{\cA}\big( S^{-1}(\cO) \big).
\end{equation}
Our method parameterizes the \emph{inverse} mapping (i.e. from observed events to aligned ones) as $S^{-1} \approx R_\omega$, so that $p_\cO(\cO | \omega) = p_\cA\big(R_\omega(\cO) \big)$.

\paragraph{Changes of volume.}
When the camera movement is a rotation, it is true that the mapping $f_t$ is a bijection on the discrete pixel set $\X$. For more general motions, even if the underlying continuous mapping is bijective, the discrete mapping may fail to be so because volume is not preserved, causing source pixels to stretch or compress so that many or no source pixels maps to a particular destination pixel $\x$. The derivation above then becomes ambiguous because the inverse image $f_t^{-1}(\x)$ may be a set of any size, including zero. The ambiguity can be resolved cleanly whenever the underlying continuous mapping is bijective by describing the entire point process in continuous spatial coordinates and correcting for the change of volume by the usual change-of-variables formula involving the determinant of the Jacobian. We leave this direction for future work. 
In our experiments with translations (non-volume-preserving mapping) we simply transform pixel coordinates of events, ignoring the change of area due to translations along the $Z$ camera axis (stretching, compressing transformations).

\end{document}